**Predicting risk of delirium from ambient noise and light information in the ICU**

**Authors:** Sabyasachi Bandyopadhyay, MS[1,6], Ahna Cecil[2], Jessica Sena, MS[3,6], Andrea Davidson[4,6], Ziyuan Guan[4,6], Subhash Nerella, MS[1,6], Jiaqing Zhang, MS[5,6], Kia Khezeli, PhD[1,6], Brooke Armfield, PhD[4,6], Azra Bihorac, MD, MS[4,6], Parisa Rashidi, PhD[1,6,*]

**Affiliations:**

[1]J. Crayton Pruitt Family Department of Biomedical Engineering, University of Florida, Gainesville, FL, USA.

[2]Department of Data Science and Business Analytics, Florida Polytechnic University, Lakeland, FL, USA.

[3]Federal University of Minas Gerais/Department of Computer Science, Belo Horizonte, Brazil.

[4]Department of Medicine, Division of Nephrology, Hypertension, and Renal Transplantation, University of Florida, Gainesville, FL, USA.

[5]Department of Electrical and Computer Engineering, University of Florida, FL, USA.

[6]Intelligent Critical Care Center (IC3), University of Florida, Gainesville, FL, USA.

*Parisa Rashidi, PhD, J. Crayton Pruitt Family Department of Biomedical Engineering, 1064 Center Drive, NEB 459, PO Box 116131, Gainesville, FL 32611. T: (352) 392-9469. Email: parisa.rashidi@bme.ufl.edu

Reprints will not be available from the authors.

**Conflicts of Interest and Source of Funding**: A.B. and P.R. were supported by NIH/NINDS R01 NS120924, NIH/NIBIB R01 EB029699, NIH/NIGMS R01 GM110240, and NIH Bridge2AI OT2OD032701, as well as the NIH/NCATS Clinical and Translational Sciences Award to the University of Florida UL1 TR000064. A.B. was supported by NIH/NIDDK R01 DK121730. P.R. was supported by NIH/NIBIB R21 EB027344 and NSF CAREER 1750192.

**Competing interests**: The authors report no conflict of interest.

**Words:** 3012

**Abstract**

**Importance:** This study reports the first deep-learning based delirium risk-assessment model for patients in the Intensive Care Unit (ICU) using only ambient noise and light information.

**Objectives:** Create a deep learning model which can prospectively classify ICU patients as delirious versus normal based on their ICU noise and light intensities.

**Design Setting and Participants:** This is a prospective cohort study in an academic hospital setting comprising 102 patients enrolled between May 2021 and September 2022.

**Main Outcomes and Measures:** Ambient light and noise intensities were measured using Thunderboard, ActiGraph sensors and an iPod with AudioTools application. These measurements were divided into daytime (0700 – 1859) and nighttime (1900 – 0659). Deep learning models were trained using this data to predict the incidence of delirium during patients' ICU stay or within 4 days of discharge. Finally, outcome scores were analyzed to evaluate each feature's importance and direction of influence.

**Results:** Daytime noise levels were significantly higher than nighttime noise levels. When using only noise features or a combination of noise and light features 1-D convolutional neural networks (CNN) achieved the strongest performance: AUC=0.77, 0.74; Sensitivity=0.60, 0.56; Specificity=0.74, 0.74; Precision=0.46, 0.40 respectively. Using only light features, Long Short-Term Memory (LSTM) networks performed best: AUC=0.80, Sensitivity=0.60, Specificity=0.77, Precision=0.37. Maximum nighttime and Minimum daytime noise levels were the strongest positive and negative predictors of delirium respectively. Nighttime light level was a stronger predictor of delirium than daytime light level. Total influence of light features outweighed that of noise features on the second and fourth day of ICU stay.

**Conclusions and Relevance:** This study shows that ambient light and noise intensities are strong predictors of long-term delirium incidence in the ICU. It reveals that daytime and nighttime environmental factors might influence delirium differently and that the importance of light and noise levels vary over the course of an ICU stay.

**Introduction**

Incidence of delirium in the Intensive Care Unit (ICU) is a significant cause of morbidity and mortality in critically ill patients[1]. Delirium is defined as an acute change in awareness and attention which develops over a short period of time and can be associated with additional cognitive disorders such as memory deficit, disorientation, and hallucination[2]. Studies suggest that delirium happens in 80% of mechanically ventilated ICU patients[3, 4] and accounts for an annual cost of 4-16 billion USD in this patient population[5]. Given the significant financial and health burdens associated with ICU delirium, its diagnosis and treatment remain imperative. Currently, delirium is diagnosed by overburdened ICU nurses using the Confusion Assessment Method ICU (CAM-ICU) questionnaire[6, 7]. The CAM-ICU contains a series of questions attempting to measure the level of attention, organization of thought, consciousness and change in mental status from baseline[8]. Due to the dependence on a baseline cognitive measurement, CAM-ICU maybe inaccurate in patients experiencing post-surgical cognitive disorder or baseline neurological disorders[9]. Furthermore, the requirement for ICU nurses to administer this test leads to sparse measurements, sometimes as low as 38%[10]. Also, CAM-ICU only identifies delirium once it has happened. Due to the vulnerability of the patient population in question, estimating the possibility of delirium onset can significantly improve their quality of care. Previous literature has mainly investigated the feasibility of predicting ICU delirium using electronic health records and vital signs[11-14]. Some studies have explored the possibility of jointly predicting level of consciousness and delirium using EEG signals from critically ill patients[15]. Despite pervasive sensing studies showing that ambient light and noise levels in ICU rooms hosting delirious patients are different from those hosting non-delirious patients[16], no study has yet developed a delirium-risk assessment model using these environmental features. Therefore, development of a delirium-risk assessment score solely using environmental factors is a critical need and a significant contribution for the ICU patient population.

Previous studies indicate that ICU rooms are subject to prolonged durations of abnormal lighting and soundscapes[17, 18]. Constant presence of artificial light, absence of windows, nightly sleep disruptions, staff conversations and sound of machinery are all contributing factors to sleep disruption and delirium in the ICU[19, 20]. Reports show that ICU noise levels are far above the WHO recommended guidelines for hospitals[21]. Some prior research has attempted to model this deleterious effect of abnormal sound levels on delirium incidence in the ICU[22].

Despite the overwhelming evidence of the impact of ambient sound and light to the incidence of delirium, no study has yet attempted to develop a delirium-risk score based on light and sound exposure in the ICU. In this study, we hypothesized that ambient light and noise intensities alone can be used to prospectively predict which ICU patients will develop delirium. We used deep neural networks to model the variegated impact of noise and light on delirium incidence. We used Shapeley Additive Explanations (SHAP) analysis to infer the relative importance of individual light and noise features towards delirium.

**Materials and Methods**

**Participants**

Participants included in this study were recruited through two federally funded clinical studies at the University of Florida (UF) Shands Hospital in Gainesville, Florida. These studies were approved by the UF Institutional Review Board and participants gave their written informed consent before study enrollment. If participants were unable to provide their informed consent, a legally authorized representative (LAR) assented on their behalf. Inclusion/Exclusion criteria: Patients were considered eligible to enroll in the study if they were greater than 18 years old, admitted to a UF intensive care unit (ICU) and expected to stay in the ICU for at least 24 hours after consenting. Patients were excluded from the study due to discharge, transfer, death within

24 hours of ICU admission or due to isolation/contact precaution requirements. Patients unable to consent were excluded from the study if they had no LAR who could consent on their behalf.

**Data Collection**

Data was collected for this study between May 2021 and September 2022 as part of two clinical investigations namely PAIN and ADAPT. In both studies, ambient data was collected for seven days or till discharge from the ICU, whichever was sooner. In the PAIN study, Actigraph GTX3+ devices were used to collect light data and iPod was used to collect noise data. In the ADAPT study, Thunderboard Sense 2 multi-sensor device was used to collect both light and noise data. Data collected using ActiGraph sensors were downloaded using the ActiLife toolbox. Noise data recorded in the iPod was collected through the AudioTools pro web application. Delirium states were calculated using the CAM, Richmond Agitation Sedation Scale (RASS) and Glasgow Coma Scale (GCS) values. These tests were administered by nurses on a daily basis.

**Training and Testing Cohorts**

In this study, we evaluated the delirium predictive power of the noise and light data separately and in combination. Correspondingly, we present three pairs of training and testing cohorts, one for each case: a) <u>Noise Cohort</u>: This contained 67 patients in the training dataset and 35 patients in the testing dataset. The training dataset comprised of 236 no-delirium samples and 79 delirium samples; while the testing dataset contained 132 no-delirium and 33 delirium samples, b) <u>Light Cohort</u>: This contained 67 patients in the training dataset and 34 patients in the testing dataset. Here, the training dataset contained 248 no-delirium days and 59 delirium days; and the testing dataset contained 147 delirium days and 23 no-delirium days, c) <u>Combined Cohort</u>: This contained 32 patients in the training dataset and 25 patients in the testing dataset. The training data comprised of 102 no-delirium days and 46 delirium days while the testing data contained 85 no-delirium and 26 delirium days. Each patient had ambient information for 1-7 ICU days, and each day had a corresponding delirium/no-delirium label.

**Preprocessing**

All data were divided into "daytime" and "night-time". These were treated as different features. Any time between 0700 to 1859 was denoted as day-time and times between 1900 to 0659 of the next day were regarded as night-time. Noise data collected through the PAIN study using the iPod and AudioTools web application had seven features: maximum noise, minimum noise, intensities greater than $99^{th}$ percentile, $90^{th}$ percentile, $50^{th}$ percentile, $10^{th}$ percentile and $1^{st}$ percentile. Therefore, noise information collected through Thunderboard multi-sensor in the ADAPT study were converted into these variables to create a combined dataset. This combined dataset was randomly sorted into training and testing datasets while ensuring that data points from the same patient do not fall into both the training and testing blocks. After division into training and testing cohorts, the information coming from PAIN and ADAPT were min-max scaled to 0-1 separately. This process is illustrated in Figure 1. The same process was followed for the light data coming from the two studies as these were also collected using different sensors. For creating the combined datasets, the training and testing datasets for noise and light were combined after their respective scaling operations. Patients who only had either the noise or the light information were removed from the final combined datasets. To feed this data into deep learning models, every sequence was zero-padded to 7 days as this was the maximum ICU length of stay in our study. Finally, multiple delirium labels for each sequence were replaced with the label which occurred more number of times. There were no ties in our dataset.

**Deep Learning Specifications**

Three different deep learning models were used in our study: a) Long Short Term Memory (LSTM), b) Gated Recurrent Units (GRU), and c) 1-dimensional convolutional neural network (1-D CNN). Each model was appended with a fully connected neural network classifier with sigmoid activation. All other layers had relu activation. All models were trained over 100 epochs with a batch size of 1 and a learning rate of 0.001. The optimized loss was binary cross entropy loss, and the optimizer was Adam. Hyper-parameters were optimized within a 3-fold cross validation setting. Several performance metrics were reported on the test datasets such as Area Under the Receiver Operating Curve (AUC), accuracy, F1-score, Precision, Sensitivity, Specificity, and negative predictive value. The test datasets were bootstrapped 100 times to generate 95% confidence intervals for each performance metric.

**SHAP Analysis**

The best model for each training condition was subjected to SHAP analysis to identify the best performing variables, and how they impacted the model score. For the combined dataset, SHAP values for noise and light were aggregated separately to investigate the relative weightage of each data modality towards the model performance.

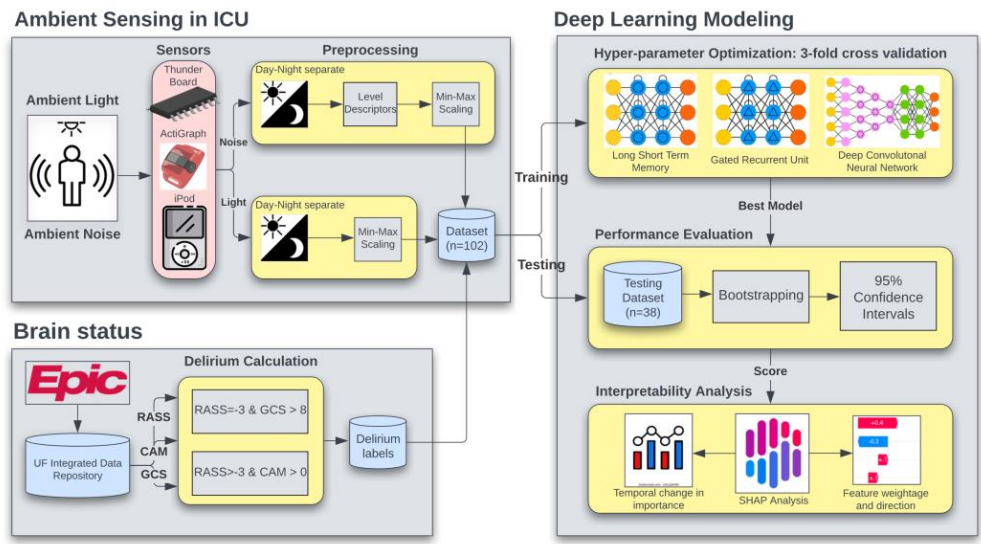

Figure 1. Conceptual Workflow of the methodology

## Results

### Participants

The participants in our study were primarily white (75.5% white, 12.7% black, 1.9% Asian), non-Hispanic (91.2% non-Hispanic, 8.8% Hispanic), biological males (66.7% male, 33.3% female), with an average age of 55.4 years (Standard Deviation; S.D=16.8 yrs). Their average length of stay in the ICU was 4.9 days (S.D=2.1 days). The distribution of days spent in the ICU resembled a bimodal distribution with two peaks at 2 days and 7 days respectively. Supplementary Fig 1 shows this distribution.

### Comparison Between Daytime and Nighttime Noise

Noise levels in the ICU during daytime and nighttime are significantly different. Table 1 shows the results of a Student's T test between maximum daytime versus maximum nighttime noise levels and between minimum daytime and minimum nighttime noise levels. Figure 2 highlights the daytime versus nighttime noise fluctuations for a single randomly selected patient from the PAIN study (P009).

Table 1. Difference between daytime and nighttime noise levels

| Noise feature | Time | Number of samples | Q1 | Median | Mean | Q3 | p-value |
|---|---|---|---|---|---|---|---|
| Lmax (dB) | Day | 7.38 X $10^6$ | 58.5 | 63.9 | 64.5 | 68.7 | 0.0 |
|  | Night | 7.41 X $10^6$ | 56.9 | 62.2 | 62.9 | 67.0 |  |
| Lmin (dB) | Day | 7.38 X $10^6$ | 52.5 | 56.5 | 58.3 | 63.5 | 0.0 |
|  | Night | 7.41 X $10^6$ | 52.2 | 55.9 | 57.8 | 63.0 |  |

Abbreviations. Lmax; Maximum noise intensity, Lmin; Minimum noise intensity, Q1; Quartile 1, Q3; Quartile 3.

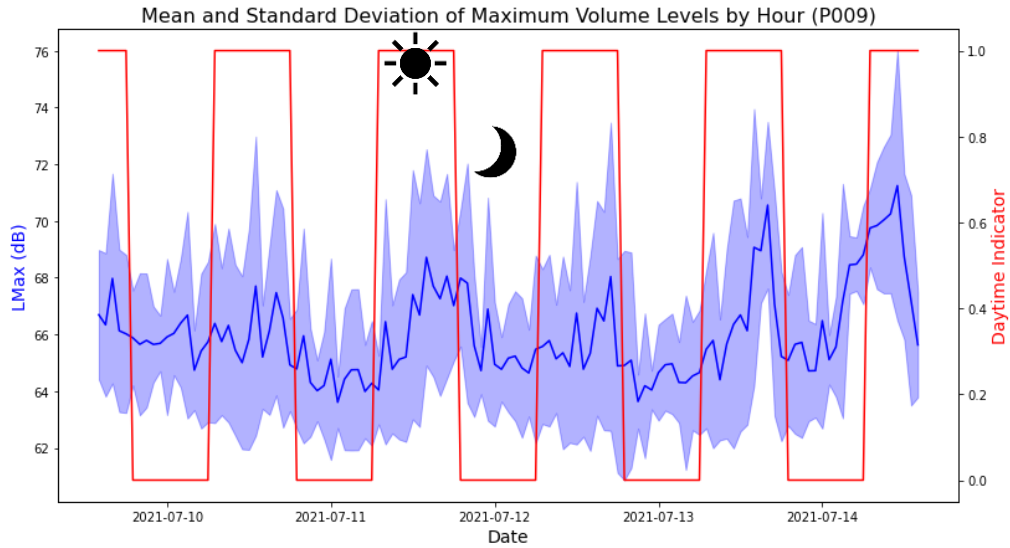

Figure 2. Mean and standard deviation of maximum noise in the ICU of a randomly selected patient (P009).

**Delirium Prediction Using Deep Learning**

Three different deep learning models capable of learning temporal sequence information were used to predict delirium from the environmental data. Table 2 shows the performance of the best model in predicting delirium during their ICU stay and over a period of 4 days from the last day in the ICU using only noise, only light and the combine data.

Table 2. Classification performance of noise and light data in predicting delirium

| Time of Delirium | Data | Method | AUC (95% C.I) | Accuracy (95% C.I) | F1-score (95% C.I) | Precision (95% C.I) | Sensitivity (95% C.I) | Specificity (95% C.I) | NPV (95% C.I) |
|---|---|---|---|---|---|---|---|---|---|
| During ICU stay or within 4 days of discharge | Noise | 1-D CNN X 2 | 0.77 (0.58 - 0.92) | 0.70 (0.51 - 0.84) | 0.50 (0.29 - 0.76) | 0.46 (0.20 - 0.76) | 0.60 (0.36 - 0.81) | 0.74 (0.55 - 0.89) | 0.86 (0.67 - 0.94) |
| | Light | LSTM | 0.80 (0.61 - 0.94) | 0.74 (0.53 - 0.91) | 0.44 (0.05 - 0.72) | 0.37 (0.04 - 0.70) | 0.60 (0.08 - 0.82) | 0.77 (0.57 - 0.93) | 0.91 (0.69 - 0.96) |
| | Combined | 1-D CNN X 2 | 0.74 (0.55 - 0.91) | 0.68 (0.54 - 0.86) | 0.43 (0.17 - 0.75) | 0.40 (0.12 - 0.76) | 0.56 (0.25 - 0.80) | 0.74 (0.57 - 0.90) | 0.83 (0.62 - 0.95) |

Abbreviations. AUC; Area Under the Curve, NPV; Negative Predictive Value

**SHAP Interpretation of Deep Learning Models**

SHAP analysis was used to investigate the order of importance and the directionality of influence of noise and light features on the deep learning models discussed above. Figure 3 shows the SHAP summary plot for the three conditions namely: predicting delirium with a) noise, b) light and c) combined. The SHAP summary plot ranks features according to their importance (topmost being the most significant predictor). Figure 3A shows that when models used only noise data, maximum nighttime noise and minimum daytime noise were the two most significant features. Furthermore, Figure 3A shows that maximum nighttime noise intensity was positively correlated with delirium while minimum daytime noise intensity was negatively correlated with delirium.

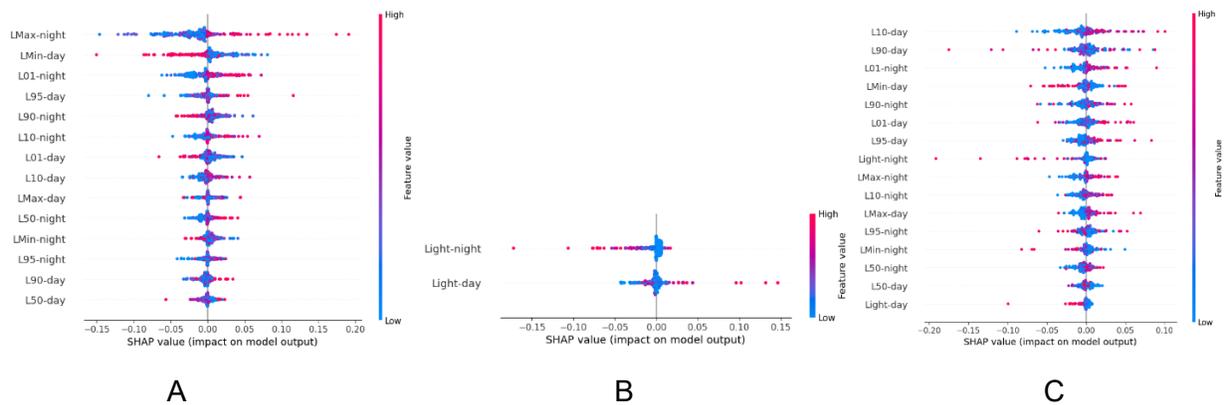

| A | B | C |

Figure 3. SHAP analysis results. A) only noise, B) only light, C) combined. Red dots indicate samples with high values of that particular feature and blue dots indicate samples with low values of that particular feature. Positive X-axis values indicate that the corresponding feature value pushed the model towards predicting 1; and vice versa.

Figure 3B reveals that light during nightly hours were protective against delirium, and light during daytime were positively associated with delirium. Figure 3C shows that when noise and light features are used together, then noise features are more significant than light features. Also, the noise features which the models found most predictive in this case were different from those in "only noise" models. In Figure 3C L10-day (10[th] percentile of daytime noise) was the most predictive feature and was positively correlated with delirium. We further investigated the relationship between noise and light features over each day of ICU stay. Figure 4A-G shows the variegated impact had by noise and light features over the course of ICU stay. Light features were more important than noise features in predicting delirium on day 2 and day 4 of ICU stay.

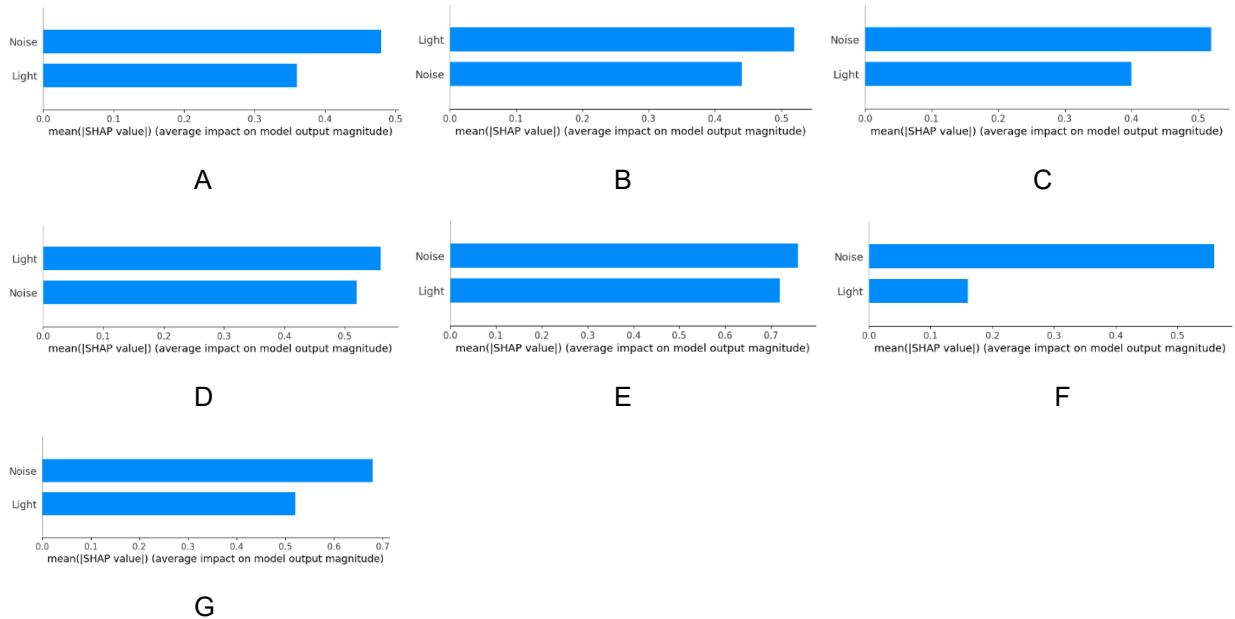

Figure 4. Difference in cumulative significance of noise versus light features over the 7 days of ICU stay. A. Day 1, B. Day 2, C. Day 3, D. Day 4, E. Day 5, F. Day 6, G. Day 7. Light > Noise only on days 2 and 4.

**Discussion**

In this study, we have used the data from two prospective single center ICU cohorts to train three temporal deep learning models to predict the risk of developing delirium inside or after discharge from the ICU. Our models show moderately strong performance in classifying patients who experienced delirium versus patients who did not. SHAP analysis revealed that maximum nighttime noise and minimum daytime noise were strong predictors of delirium but differed in their directionality of effect. While maximum nighttime noise was a positive predictor, minimum daytime noise was a negative predictor of delirium. SHAP analysis also found that higher nighttime noise intensities corresponded to less prevalence of delirium while higher daytime noise intensities corresponded to more prevalence of delirium in our dataset. Combining light and noise intensities resulted in reduced performance over a range of metrics.

To our knowledge, this study is the first of its kind to investigate the feasibility of predicting delirium incidence in the ICU by using only environmental factors such as ambient light and noise. Although there is considerable prior research which indicates that the abnormal noise and light inside ICU rooms contribute to circadian desynchrony which causes delirium no study has yet attempted to construct a delirium-risk score based solely on these environmental factors. Furthermore, this study represents a growing body of data where researchers used pervasive sensing in the ICU to improve the predictability of adverse outcomes. Additionally, our study offers some unique strengths. The use of deep learning allowed us to use environmental factors in a non-linear manner to predict delirium. This is evident in the case of light features where SHAP analysis revealed the directionality of the nighttime light to be ambiguous, yet the model retained it as the most significant feature. In the combined cohort, L90-day significantly predicted delirium without having a strict directionality. This meant that the decision boundaries found by our model were non-linear. Furthermore, our study did not treat the deep learning models as black boxes. SHAP analysis revealed the relative importance of different features, how they impacted the model output and the relative impact of combined noise and light features on different days of ICU stay. The fluctuating importance of light features indicated a possible relationship between nighttime light intensities and patient care schedules. This study is also the first to compare the importance of ICU noise and light intensities together and in isolation.

Our study has some weaknesses. The presence of only 102 patients in our overall cohort weakened the robustness of our deep learning models. This was exacerbated in the combined cohort where the training data only comprised of 32 patients. This prevented us from training models with large number of parameters due to possibility of overfitting. The loss of samples in the combined cohort was a result of using different sensors to collect light and noise data in the PAIN study. The periods of measurement of light and noise data were often not the same and hence these samples had to be excluded from the combined cohort. Another disadvantage of using multiple sensors was the possible presence of batch effect. In the noise data, we countered this by computing the different features collected in the PAIN study from the Thunderboard data (ADAPT study). The distributions of noise intensities in both studies were Gaussian (Supplementary Figure 2A-B). However, the light intensity distributions were dissimilar between the ActiGraph (PAIN) and the Thunderboard (ADAPT) studies (Supplementary Figure 2C-D). This could be an influencing factor behind the apparent loss of directionality in the light features.

Future work will revolve around collecting more environmental data using a single multi-sensor platform. We will use this data to further improve the deep models and test them on larger independent validation datasets. The temporally fluctuating importance of light features also warrants further investigation. We think that this might be caused by nightly disruptions owing to patient care. It is possible that the significance of nighttime light intensities as a delirium predictor on day 2 and 4 of ICU stay is really a reflection of a loss of circadian rhythm caused by scheduled nighttime patient care on these days. Therefore, we will incorporate sleep quality questionnaires after each night spent in the ICU into our analytic models. We will use state-of-the-art Natural Language Processing architectures such as Transformers to model this information.

**Conclusions**

Deep Learning based time-series modeling of environmental data could develop classifiers with good performance for identifying patients who prospectively developed delirium. Further investigation of the relative importance of different environmental factors revealed that while noise remains an important predictor of delirium throughout the ICU stay, light becomes more important than noise on certain nights of ICU stay.

**Acknowledgments**

The authors thank all patients and their families and research coordinators for participating in this study.

**Author Contributions**

PR conceived the original idea for the study and sought and obtained funding. SB had full access to all the data in the study and take responsibility for the integrity of the data and the accuracy of the data analysis. AC, JS, JZ had partial access to the data and are responsible for some analyses. AD, ZG, SN were responsible for data collection. Interpretation of data: All authors. The article was written by SB with input from all coauthors. All authors participated in critically revising the manuscript for important intellectual content and gave final approval of the version to be published. AB, PR and KK served as senior authors.

# Supplementary Information

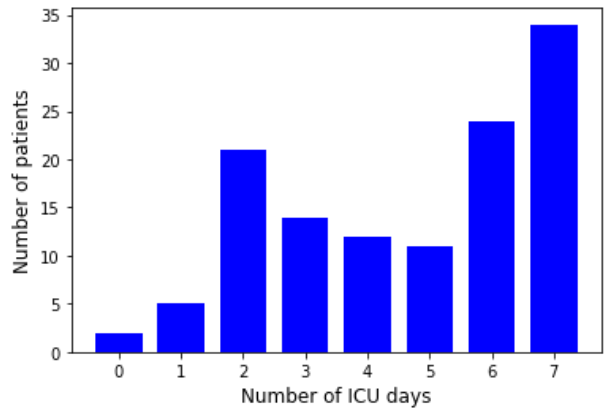

Supplementary Fig 1. Distribution of number of days spent in the ICU. The distribution is bimodal with peaks at 2 days and 7 days.

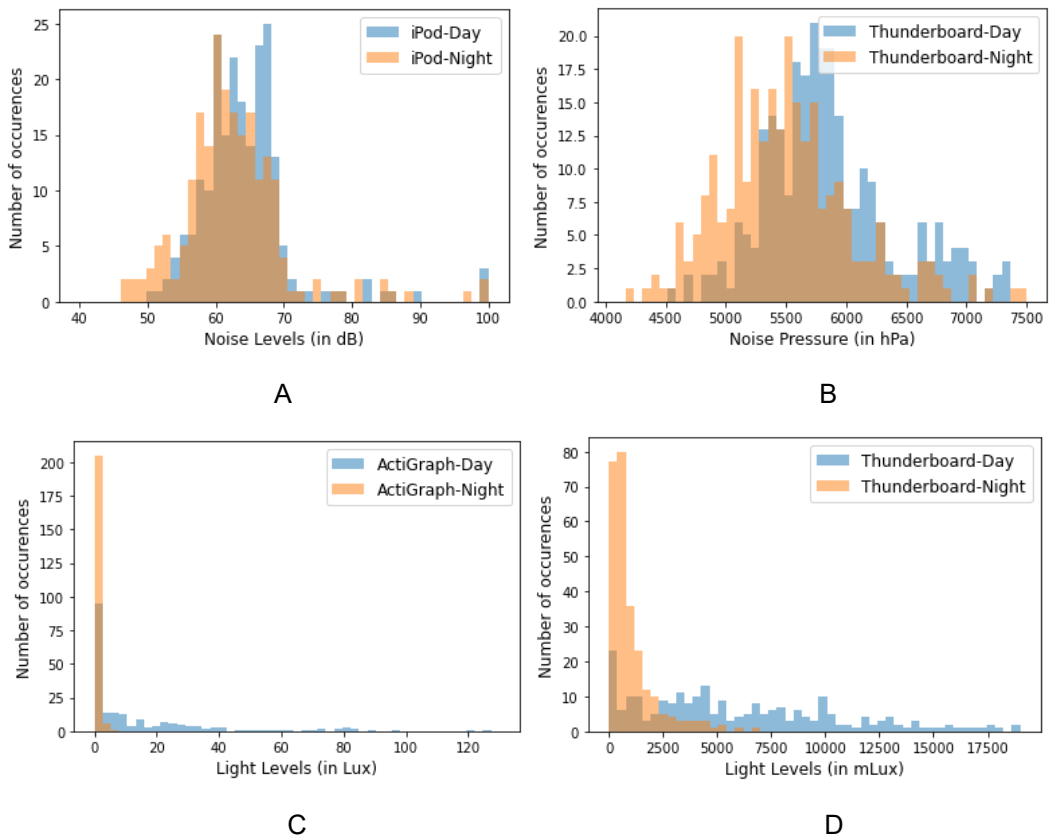

Supplementary Fig 2. Distribution of maximum noise levels and light intensities collected through different sensors. A. Noise levels collected through iPod. B. Noise levels collected through Thunderboard. C. Light levels collected using ActiGraph. D. Light levels collected using Thunderboard. Noise intensities are Gaussian irrespective of the sensor, but light intensities are positively skewed and different between ActiGraph and Thunderboard sensors.